\documentclass{article}
\usepackage{amsmath,graphicx,mlspconf}
\usepackage{xcolor}
\newcommand{\ie}{\textit{i}.\textit{e}.,\ }
\newcommand{\eg}{\textit{e}.\textit{g}.,\ }
\usepackage[pagebackref,breaklinks,colorlinks]{hyperref}

 \usepackage{booktabs}
 \usepackage{multirow}
\usepackage[capitalize]{cleveref}
\crefname{section}{Sec.}{Secs.}
\Crefname{section}{Section}{Sections}
\Crefname{table}{Table}{Tables}
\crefname{table}{Tab.}{Tabs.}

\title{Quick Bypass Mechanism of Zero-Shot Diffusion-Based \\ Image Restoration}
\name{Yu-Shan Tai, An-Yeu (Andy) Wu}
\address{\textit{Graduate Institute of Electrical Engineering}\\\textit{National Taiwan University}\\Taipei, Taiwan\\ clover@access.ee.ntu.edu.tw,  andywu@ntu.edu.tw}

\begin{document}

\maketitle

\begin{abstract}
    Recent advancements in diffusion models have demonstrated remarkable success in various image generation tasks. Building upon these achievements, diffusion models have also been effectively adapted to image restoration tasks, \eg super-resolution and deblurring, aiming to recover high-quality images from degraded inputs. Although existing zero-shot approaches enable pretrained diffusion models to perform restoration tasks without additional fine-tuning, these methods often suffer from prolonged iteration times in the denoising process. To address this limitation, we propose a Quick Bypass Mechanism (QBM), a strategy that significantly accelerates the denoising process by initializing from an intermediate approximation, effectively bypassing early denoising steps. Furthermore, recognizing that approximation may introduce inconsistencies, we introduce a Revised Reverse Process (RRP), which adjusts the weighting of random noise to enhance the stochasticity and mitigate potential disharmony. We validate proposed methods on ImageNet-1K and CelebA-HQ across multiple image restoration tasks, \eg super-resolution, deblurring, and compressed sensing. Our experimental results show that the proposed methods can effectively accelerate existing methods while maintaining original performance.

\end{abstract}

\section{Introduction}

\label{Introduction}
Diffusion models (DMs) have demonstrated significant advances in image generation \cite{ddpm, dm_high}, video generation \cite{video, video_imagen, videoDM}, and 3D point cloud generation \cite{point_cloud, 3d_conditional}. These models progressively add noise to input data until it conforms to a Gaussian distribution, and subsequently learn to denoise, restoring real data from the sampled noise.
Recently, DMs have been further applied to image restoration tasks, such as super-resolution and deblurring \cite{ddnm, denoising, dm_srdiff}, whose objective is to enhance image quality from a degraded input.

\begin{figure}[t]
    \centering
    \includegraphics[width=1\columnwidth]{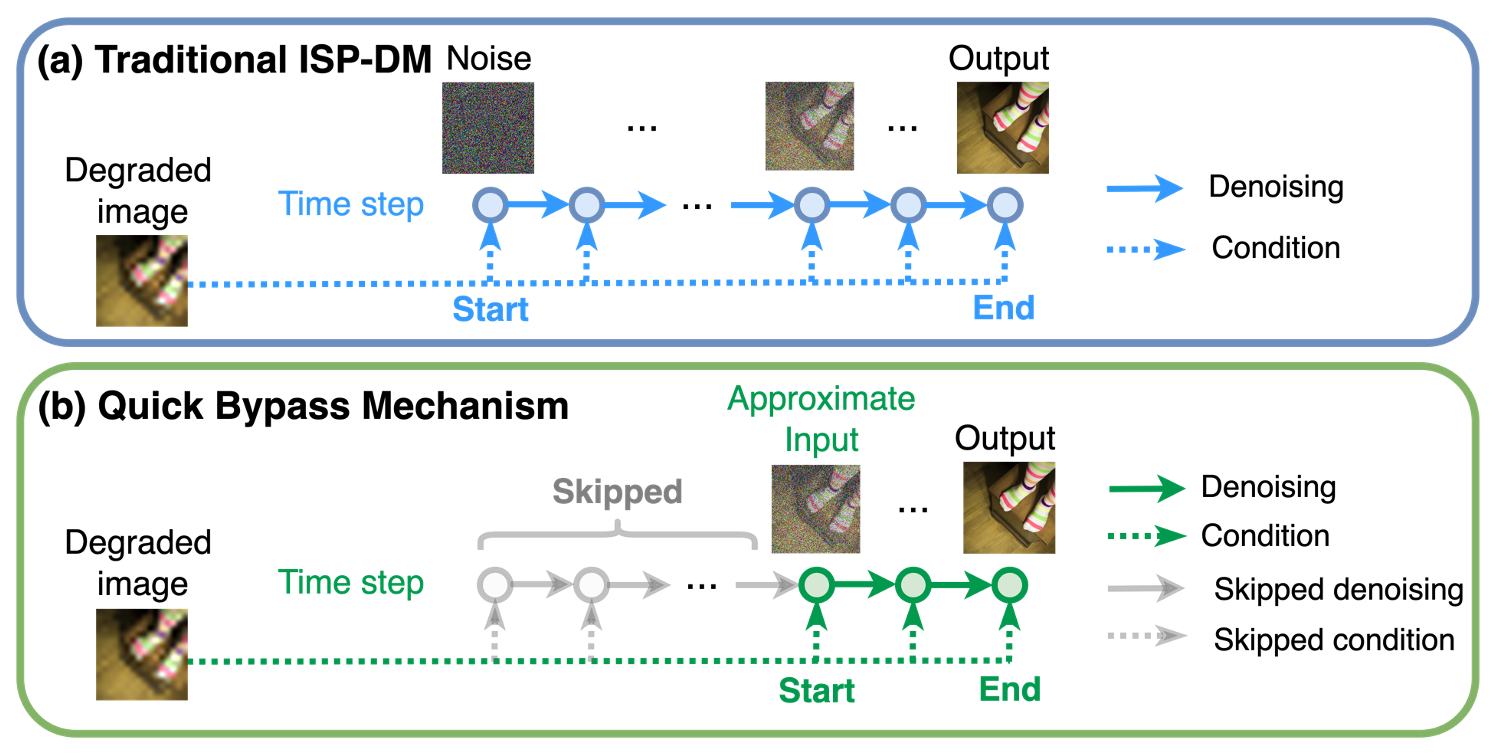}
    \caption{\textbf{Denoising process of image restoration.} (a) Original zero-shot diffusion model, \textit{e.g.,} DDNM \cite{ddnm}, start denoising from pure noise. (b) Our proposed method uses an approximate input to save early-stage steps.}
   \label{fig:overview}
\end{figure}

However, due to the multi-step denoising required inherent in DMs, they are limited in real-time applications, such as online super-resolution streaming or self-driving cars. 
Although several methods have been proposed to accelerate DMs, most of these focus on unconditional generation tasks \cite{ptq4dm, accelerating, pseudo}. Unlike generation tasks, which require denoising to begin from pure noise, image restoration tasks involve a degraded input. During denoising, early stages target reconstructions of structural or coarse features, many of which are already present in degraded images. This suggests that the degraded input could be leveraged to create an approximate input, allowing denoising to start from an intermediate point, rather than from pure noise, as shown in \cref{fig:overview}.

To identify the optimal point to begin denoising, we must determine the most suitable time step for the bypass. That is, the approximate input should be sufficiently close to the original trajectory derived from the full multi-step denoising process. We need to select a time step with acceptable deviation while also maximizing the potential acceleration it offers. Additionally, the disharmony introduced by the bypass must be compensated for, ensuring minimal impact on the overall denoising process.

To address the challenges outlined above, we propose a novel approach called Quick Bypass Mechanism for Zero-Shot Diffusion-Based Image Restoration. Our method introduces a Quick Bypass Mechanism (QBM) to identify the optimal time step to initiate denoising. Through a detailed analysis of the relationship between approximate inputs and original trajectories, we establish two key criteria that ensure their similarity.
To further mitigate the potential disharmony caused by the bypass strategy, we propose a Revised Reverse Process (RRP). This process enhances the stochastic nature of the denoising procedure by increasing the weighting of random noise, thereby improving stability throughout the denoising process.

We evaluate the proposed method on ImageNet-1K \cite{imagenet} and CelebA-HQ \cite{celeba} across three image restoration tasks, \eg super-resolution, deblurring, and compressed sensing. Our experimental results demonstrate that our approach achieves performance comparable to or even surpassing the original restoration model, while requiring only 5\%$\sim$59\% of the original time steps. This highlights the effectiveness of our method in accelerating diffusion-based image restoration.

\section{Related Work}
\label{Related_Work}

\subsection{Diffusion-Based Image Restoration}
\textbf{Diffusion Model. } \label{sec:DM}
    Diffusion models (DMs) \cite{ddpm} generate images using a Markov chain. 
    Given a real data distribution $x_{0} \sim q(x_{0})$, the diffusion process gradually adds Gaussian noise with a predefined variance schedule $\beta_{t} \in (0,1)$ to produce an intermediate input sequence $x_{1}, ..., x_{T}$ with the total number of steps, $T$.
    The intermediate inputs $x_{t}$ from arbitrary $t$ can be directly derived from $x_{0}$ via the forward process:
    \begin{equation} \label{eq:forward}
        x_{t} = \sqrt{\bar{\alpha}_{t}} x_{0} + \sqrt{1 - \bar{\alpha}_{t}} \epsilon, 
    \end{equation}
    where ${\alpha}_{t}=1-\beta_{t}$, $\bar{\alpha}_{t}=\prod_{t=1}^{T}\alpha_{t}$, and $\epsilon \in \mathcal{N}(0, \mathbf{I})$.
    Conversely, the denoising process aims to remove noise from a sampled noise input $x_{T}\in \mathcal{N}(0,\mathbf{I})$ to generate the final output $x_{0}$.
    Since $q(x_{{t-1}}|x_{{t}})$ is intractable, DM uses a model parameterized by $\theta$ to approximate the conditional distribution. The estimation of the final images at time $t$ can be derived from:
    \begin{equation} \label{eq:ddim_predict}
        x_{0|t} = \frac{x_{t} - \sqrt{1 - \bar{\alpha}_{t}} \, \epsilon_\theta(x_{t}, t)}{\sqrt{\bar{\alpha}_{t}}}.
    \end{equation}
    Consequently, the next-step input $x_{t-1}$ is determined by a hyperparameter $\eta$ that controls the trade-off between random and estimated noise, as given by the following equation:
    \begin{equation} \label{eq:ddim_next}
        x_{t-1} =  \sqrt{\bar{\alpha}_{t}} x_{0|t} + \sqrt{1 - \bar{\alpha}_{t}} (\eta\cdot\epsilon+\sqrt{1-\eta^2}\cdot\epsilon_\theta(x_{t}, t)).
    \end{equation}
    
\textbf{DMs for Image Restoration. } 
    Given the remarkable performance achieved in generative tasks, DMs have been further extended to image restoration tasks.  These tasks are essential for improving the quality of a degraded image $y$, typically framed as linear inverse problems with a common formulation of $y = A(x_0)$, where $A$ represents a known linear operator and $x_0$ is the high-quality (HQ) image. 
    
    Some researchers introduce zero-shot methods that leverage pretrained DMs as generative priors without additional training or network modifications.  
    A common strategy is to decompose the image into different spaces to facilitate image restoration. For example, DDRM \cite{ddrm} employs singular value decomposition (SVD) of the degradation operator and performs denoising in the spectral space. DDNM \cite{ddnm}, on the other hand, leverages a DM to generate only the null-space components. Then, they directly utilize the degraded input and apply inverse degradation, \( A^{\dagger}y \), to replace the range-space components \(A^{\dagger}A (x_{0|t})\):
    \begin{equation}
        \hat{x}_{0|t} = A^{\dagger}y + \left( I - A^{\dagger}A\right)x_{0|t}.
        \label{eq:ddnm}
    \end{equation}
    
    While these methods successfully adapt DMs to image restoration, they retain the generative process of denoising from pure noise, which hinders their applicability in real-time scenarios. Thus, it is essential to develop specialized methods to accelerate diffusion-based image restoration.

\subsection{Acceleration for Diffusion Models}
\textbf{Knowledge Distillation. } 
    Recent studies have leveraged knowledge distillation to enable single-step or few-step DMs to produce results comparable to those of multi-step models \cite{consistency, consistency_latent}. However, these methods often require substantial training resources. For instance, Consistency Models \cite{consistency} necessitate 800k iterations for training. Additionally, even achieving generation tasks within a single step, image restoration tasks still demand multiple steps to achieve high-quality outputs.
    
\textbf{Training-free Sampler. }
    To alleviate the significant training overhead, some researchers have proposed training-free samplers that adjust the operations performed at each step during denoising to reduce the required steps. For example, DDIM \cite{ddim} utilizes a non-Markovian process, employing a uniform sub-sequence of DDPM \cite{ddpm} to effectively minimize the number of steps. 
    Furthermore, DPM-Solver \cite{dpm} and DPM-Solver++ \cite{dpm_solver++} introduce a specialized high-order solver that analytically computes the linear components of diffusion ordinary differential equations (ODEs).
    However, while these samplers are general, they lack the flexibility to adapt to different data distributions or specific tasks.
    
    In contrast to these approaches, we propose a training-free acceleration method tailored to image restoration tasks. Our novelty lies in introducing a bypass mechanism that begins the denoising process from an intermediate state to avoid starting from pure noise.

\section{Approach}
\label{Approach}
To accelerate diffusion-based image restoration, we propose the \textbf{Quick Bypass Mechanism (QBM)}, which starts denoising from an intermediate step using an approximate input instead of pure noise. Furthermore, to address the discrepancy induced by the approximation, we introduce the \textbf{Revised Reverse Process (RRP)} to increase the weighting of random noise to enhance stochasticity.
\subsection{Quick Bypass Mechanism (QBM)}
\begin{figure}[t]
    \centering
    \includegraphics[width=1\columnwidth]{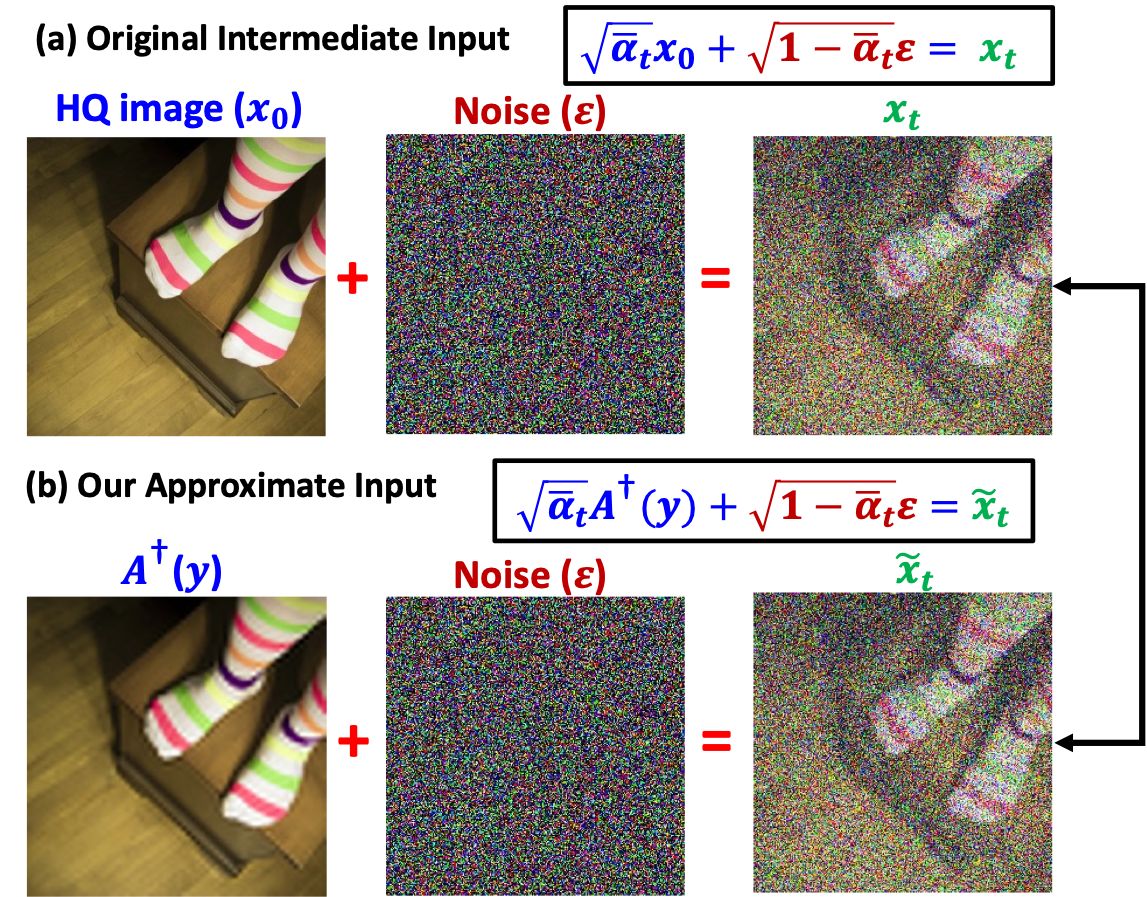}
    \caption{\textbf{Our approximate input}. To enable the denoising process to begin from an intermediate step, our Quick Bypass Mechanism (QBM) identifies an initial time step $t$ to construct an approximate input ($\tilde{x}_t$ in (b)) that closely resembles the original intermediate input ($x_t$ in (a)), enabling diffusion models to start from $t$ rather than a large $T$.}
   \label{fig:idea}
\end{figure}

   In our proposed Quick Bypass Mechanism (QBM), we aim to construct an approximate input at an intermediate time step $t$ to facilitate a faster denoising process. Existing diffusion-based image restoration \cite{ddnm, ddrm} follows the same denoising procedure as in generative tasks, which typically begin with pure noise. However, unlike generative tasks, image restoration tasks receive a degraded image $y$ as input, which provides additional information, such as structural or coarse features. Since the early stages of the denoising process primarily focus on reconstructing high-level features, we propose that extracting the information already present in the degraded input can bypass this stage, thus accelerating the overall process.

   \subsubsection{Approximate Input} To construct the approximate input, we propose replacing the HQ image $x_{0}$ in \cref{eq:forward} with the degraded image $y$, processed by pseudo-inverse operation of the degradation function $A$, \ie $A^{\dagger}(y)$. Thus, the approximate input $\tilde{x}_t$ at time step $t$ can be constructed as: 
   \begin{equation} 
   \tilde{x}_{t} = \sqrt{\bar{\alpha}_{t}} A^{\dagger}(y) + \sqrt{(1-\bar{\alpha}_{t})} \epsilon. 
   \label{eq:AS}
   \end{equation} 

   \subsubsection{Time Step for Bypass} \label{sec:intiial_t}
   Then, the next challenge is to determine the appropriate time step to start denoising, as illustrated in \Cref{fig:idea}. Specifically, we need to identify a time step $t$ where the approximate input ($\tilde{x}_t$) closely resembles the original intermediate input ($x_t$). Intuitively, since intermediate inputs with larger $t$ are noisier, larger values of $t$ ensure better similarity. However, this also means more steps are required in the process. Therefore, it is crucial to find the smallest $t$ that maintains good consistency with the original input.

    To find the optimal time step $t$ to begin denoising, we need to analyze the difference between the approximate input \(\tilde{x}_{t}\) and the original intermediate input \(x_{t}\). The original \(x_{t}\) can be represented as:
    
    \begin{equation}
    x_{t} = \sqrt{\bar{\alpha}_{t}}x_{0} + \sqrt{1-\bar{\alpha}_{t}} \epsilon
    \end{equation}
        \begin{equation}
     = \sqrt{\bar{\alpha}_{t}} \left( A^{\dagger}(y) + (x_{0} - A^{\dagger}(y)) \right) + \sqrt{1-\bar{\alpha}_{t}} \epsilon
    \end{equation}
    \begin{equation}
     = \sqrt{\bar{\alpha}_{t}} A^{\dagger}(y) + \sqrt{\bar{\alpha}_{t}} (x_{0} - A^{\dagger}(y)) + \sqrt{1-\bar{\alpha}_{t}} \epsilon.
    \end{equation}
    To ensure similarity between \(x_{t}\) and \(\tilde{x}_{t}\) as described in \cref{eq:AS}, the discrepancy $\sqrt{\bar{\alpha}_{t}} (x_{0} - A^{\dagger}(y))$ should be absorbed into the noise term. That is, the noise must be large enough to cover the discrepancy, which can be formulated as:
    \begin{equation}
     \sqrt{\bar{\alpha}_{t}} (x_{0} - A^{\dagger}(y)) + \sqrt{1-\bar{\alpha}_{t}} \epsilon \sim  \sqrt{1-\bar{\alpha}_{t}} \epsilon.
     \label{eq:condtion}
    \end{equation}
    To find the minimum time step $t$ that satisfies the relationship in \cref{eq:condtion}, we propose using two criteria for evaluation.
    
    \textbf{Criteria 1}: $\sqrt{\bar{\alpha}_{t}} (x_{0} - A^{\dagger}(y)) + \sqrt{1-\bar{\alpha}_{t}} \epsilon$ should follow a Gaussian distribution. Since $\sqrt{1-\bar{\alpha}_{t}} \epsilon$ is Gaussian noise, we need to ensure that the approximation also follows this distribution. Based on  D'Agostino and Pearson's test \cite{d1973tests, diagostino1971omnibus}, we use skewness and kurtosis of a distribution to determine whether it conforms to a Gaussian distribution.
    
    \textbf{Criteria} 2: The standard deviation of $\sqrt{\bar{\alpha}_{t}} (x_{0} - A^{\dagger}(y)) + \sqrt{1-\bar{\alpha}_{t}} \epsilon$ should be close to that of $\sqrt{1-\bar{\alpha}_{t}} \epsilon$. Since the noise at each time step of the denoising process has a different standard deviation (controlled by $\beta_t$ as mentioned in \cref{sec:DM}), we need to ensure that the noise maintains the expected characteristics for this step. Specifically, we set a threshold $k$ to examine whether their standard deviations are close enough:
    \begin{equation}
     |std(\sqrt{\bar{\alpha}_{t}} (x_{0} - A^{\dagger}(y)) + \sqrt{1-\bar{\alpha}_{t}} \epsilon)-std(\sqrt{1-\bar{\alpha}_{t}} \epsilon)|<k.
     \label{eq:condtion2}
    \end{equation}

    Using these two criteria, we offline analyze the minimum $t$ that satisfies both from a calibration set and use the average result as the initial denoising step.  The search process is efficient and fast, as it does not require model inference.

\subsection{Revised Reverse Process (RRP)}
    From here, we have introduced QBM to utilize an approximate input to replace an intermediate input derived from multiple time steps. However, this approximation inevitably introduces some disharmony, as it deviates from the original intermediate input. To mitigate this disharmony and ensure stable denoising, we propose a Revised Reverse Process (RRP) to increase the stochasticity during the denoising process.
    
    Specifically, we modify the weighting of random noise $\epsilon$ and estimated noise $\epsilon_\theta(x_t, t)$ in \cref{eq:ddim_next}.
    Typically, in generative tasks, DDIM \cite{ddim} sets the parameter $\eta$ to 0 for a deterministic trajectory. In contrast, for image restoration tasks, DDNM \cite{ddnm} applies $\eta = 0.85$ to reduce the disharmony induced by \cref{eq:ddnm}. However, in image restoration, where $x_{0|t}$ in \cref{eq:ddim_next} is replaced by $\hat{x}_{0|t}$ in \cref{eq:ddnm}, the estimated noise $\epsilon_\theta(x_t, t)$ becomes less meaningful. This is because the image $x_{0|t}$ has been manually modified, meaning the noise no longer accurately represents the noise along the trajectory.
    Therefore, instead of combining random noise with the estimated noise, we propose relying solely on random noise by setting $\eta$ to 1. This approach maximizes the potential to alleviate disharmony caused by the approximation.

\section{Experiment}
\label{Experiment}
\begin{table}[t]
    \caption{\textbf{Quantitative results on ImageNet-1K (256$\times$256).} We evaluate the images using PSNR and SSIM across three image restoration tasks. Our proposed QBM+RRP achieves superior performance compared to the 100-step DDNM \cite{ddnm}, while requiring fewer steps.}
\scalebox{0.9}{\begin{tabular}{ccccc}
\toprule
Task                                & Method             & \# of steps & PSNR           & SSIM           \\ \midrule
\multirow{5}{*}{Super-resolution}   & DDNM   \cite{ddnm}             & 100         & 27.45          & 0.889          \\
                                    & DDNM   \cite{ddnm}             & 33          & 27.23          & 0.886          \\
                                    & QBM only             & 33          & 27.33          & 0.888          \\
                                    & RRP only             & 33          & 27.43          & 0.890          \\
                                    & \textbf{QBM+RRP} & \textbf{33} & \textbf{27.52} & \textbf{0.892} \\\midrule
\multirow{5}{*}{Deblurring}         & DDNM    \cite{ddnm}            & 100         & 45.13          & 0.995          \\
                                    & DDNM   \cite{ddnm}             & 10          & 42.02          & 0.992          \\
                                    & QBM only             & 10          & 45.02          & 0.995          \\
                                    & RRP only             & 10          & 41.86          & 0.992          \\
                                    & \textbf{QBM+RRP} & \textbf{10} & \textbf{45.89} & \textbf{0.996} \\\midrule
\multirow{5}{*}{Compressed sensing} & DDNM   \cite{ddnm}             & 100         & 21.74          & 0.702          \\
                                    & DDNM  \cite{ddnm}              & 57          & 19.55          & 0.619          \\
                                    & QBM only             & 57          & 20.31          & 0.667          \\
                                    & RRP only             & 57          & 21.52         & 0.715          \\
                                    & \textbf{QBM+RRP} & \textbf{57} & \textbf{22.84} & \textbf{0.770} \\ \bottomrule
\end{tabular}} \label{tab:imagenet}
\end{table}

\subsection{Experimental Setup}
Our proposed method is a plug-and-play framework that can be seamlessly integrated with existing zero-shot DMs for image restoration tasks. In the following experiments, we implement our method based on DDNM \cite{ddnm}. We evaluate performance on two datasets: ImageNet-1K \cite{imagenet} and CelebA-HQ \cite{celeba}, both at a resolution of 256$\times$256.
For ImageNet-1K, we sample 1,000 images from the training set for calibration and use the official validation set for testing. For CelebA-HQ, we randomly select 1,000 images from the total dataset of 30,000 images for calibration and another 1,000 images for testing. The threshold $k$ used in RRP is set as 0.001. We perform experiments on various image restoration tasks, including super-resolution, deblurring, and compressed sensing.
The quality of the restored images is assessed using Peak Signal-to-Noise Ratio (PSNR) and Structural Similarity Index Measure (SSIM) for a comprehensive evaluation of image quality.
\subsection{Quantitative Results} \label{sec:quanti}

\begin{table}[t]
    \caption{\textbf{Quantitative results on CelebA-HQ (256$\times$256).} We evaluate PSNR and SSIM across three tasks. Our QBM+RRP outperforms DDNM \cite{ddnm} with the same number of steps and approaches the result of 100-step DDNM \cite{ddnm}.}
\scalebox{0.9}{\begin{tabular}{ccccc}
\toprule
Task                                & Method             & \# of steps & PSNR           & SSIM           \\ \hline
\multirow{5}{*}{Super-resolution}   & DDNM \cite{ddnm}               & 100         & 31.71          & 0.950          \\
                                    & DDNM \cite{ddnm}               & 23          & 31.70          & 0.950          \\
                                    & QBM only             & 23          & 31.72         & 0.950          \\
                                    & RRP only             & 23          & 31.70        & 0.951        \\
                                    & \textbf{QBM+RRP} & \textbf{23} & \textbf{32.70} & \textbf{0.959} \\\midrule
\multirow{5}{*}{Deblurring}         & DDNM   \cite{ddnm}             & 100         & 55.82          & 1.000          \\
                                    & DDNM  \cite{ddnm}              & 5           & 48.72          & 0.998          \\
                                    & QBM only             & 5          & 55.02         & 1.000          \\
                                    & RRP only             & 5          & 48.62        & 0.998        \\
                                    & \textbf{QBM+RRP} & \textbf{5}  & \textbf{55.49} & \textbf{1.000} \\\midrule
\multirow{5}{*}{Compressed sensing} & DDNM   \cite{ddnm}             & 100         & 27.68          & 0.891          \\
                                    & DDNM  \cite{ddnm}              & 59          & 25.70          & 0.855          \\
                                    & QBM only             & 59          & 25.38         & 0.859          \\
                                    & RRP only             & 59          & 26.56        & 0.870        \\
                                    & \textbf{QBM+RRP} & \textbf{59} & \textbf{27.46} & \textbf{0.892} \\ \bottomrule
\end{tabular}} \label{tab:celeba}
\end{table}

In \cref{tab:imagenet} and \cref{tab:celeba}, we present the PSNR and SSIM of DDNM \cite{ddnm} and our methods on ImageNet-1K \cite{imagenet} and CelebA-HQ \cite{celeba}, respectively. We show DDNM \cite{ddnm} with 100 steps as a baseline, while other settings are run with the same number of steps as determined by QBM (\cref{sec:intiial_t}). The results show that both QBM and RRP individually improve performance, and combining them (QBM+RRP) yields the best results. Our method outperforms DDNM \cite{ddnm} across all tasks with the same number of steps and even surpasses the 100-step DDNM \cite{ddnm} in most cases.

\begin{figure*}[t]
    \centering
    \includegraphics[width=0.9\linewidth]{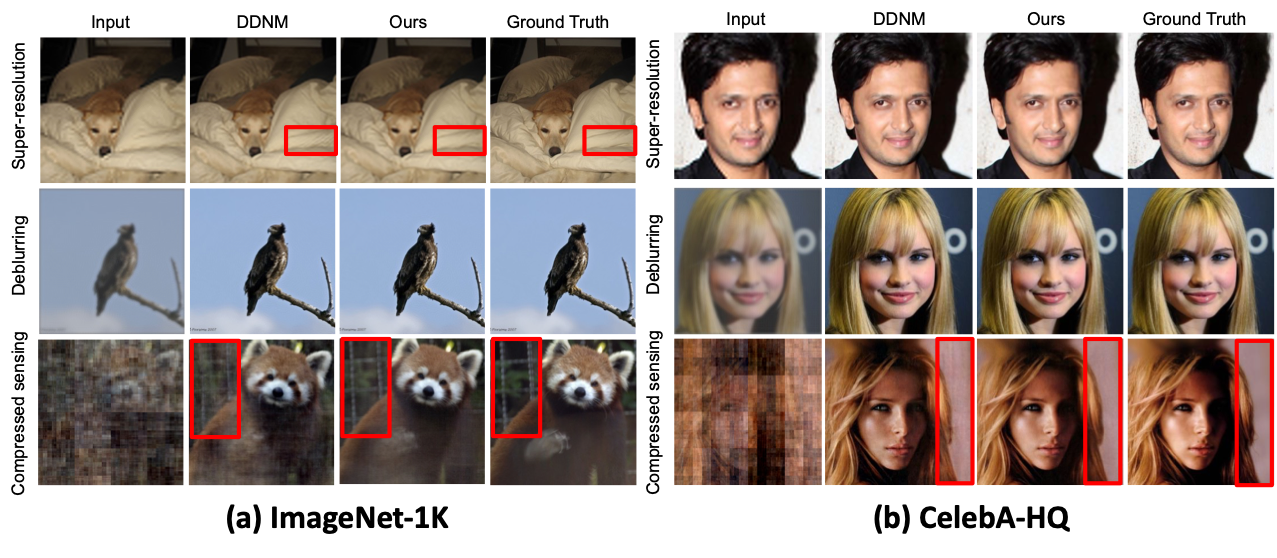}
    \caption{\textbf{Qualitative Results.} We show the outputs of our method and DDNM \cite{ddnm} with corresponding degraded inputs and ground truth images from (a) ImageNet-1k \cite{imagenet} and (b) CelebA-HQ \cite{celeba}. Our results demonstrate a superior ability to restore fine-grained image features.}
    \label{fig:visual}
\end{figure*}

We notice that the number of steps derived from QBM varies across tasks. Since the degradation operations differ, the amount of noise required to cover the discrepancy also differs. Generally, a larger discrepancy requires more steps. We observe that deblurring requires the fewest steps in both datasets. This is because deblurring introduces Gaussian noise to the HQ image, making it more similar to the diffusion process in DMs and thus easier compared to other tasks.

\subsection{Qualitative Results}

In \cref{fig:visual}, we visualize the outputs of DDNM \cite{ddnm} and our method with the same number of steps on ImageNet-1K \cite{imagenet} and CelebA-HQ \cite{celeba}. Our method demonstrates better image quality restoration. For instance, DDNM \cite{ddnm} may produce unnatural wrinkles on sheets in (a) super-resolution or defects in backgrounds in (a)(b) compressed sensing. Deblurring is a relatively simple task, making the differences between methods harder to discern.

\subsection{Gaussian Distribution Analysis}

\begin{figure}[h!]
    \centering
    \includegraphics[width=1\linewidth]{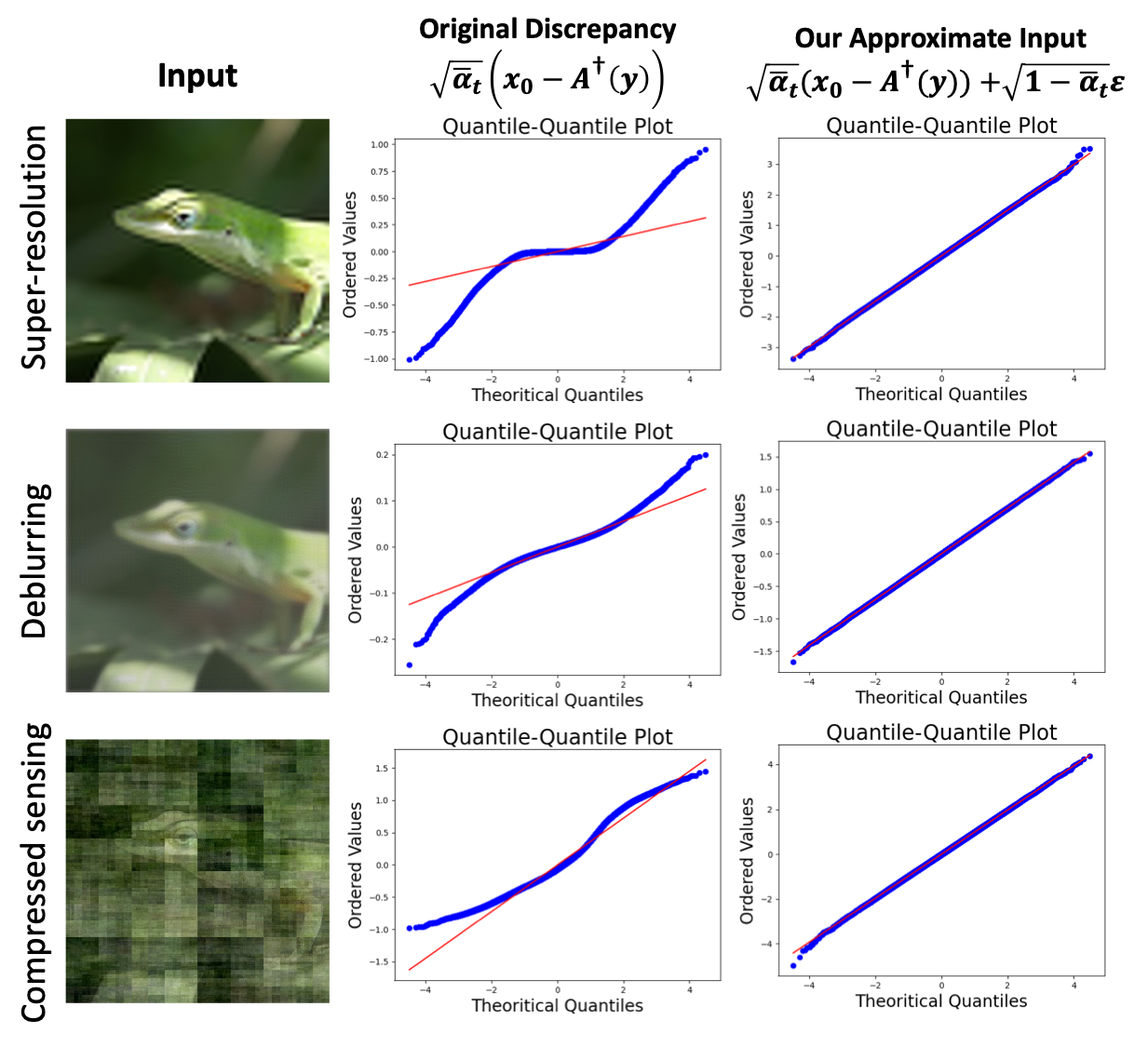}
    \caption{\textbf{Gaussian Distribution Analysis.} We show the quantile-quantile plots of the original discrepancy and our approximate input. If the line is closer to a diagonal, the distribution is closer to Gaussian. The result shows that our approximate input satisfies Criteria 1 mentioned in \cref{sec:intiial_t}.}
    \label{fig:qq}
\end{figure}

In \cref{fig:qq}, we show the quantile-quantile plots (Q-Q plots) of the original discrepancy and our approximate input. A Q-Q plot is a graphical tool to assess if a set of data plausibly follows a theoretical distribution. If the data closely matches the distribution, the points on the Q-Q plot will align along a diagonal line. We observe that the original discrepancy $\sqrt{\bar{\alpha}_{t}} (x_{0} - A^{\dagger}(y))$ significantly deviates from a Gaussian distribution. However, after introducing sufficient random noise $\sqrt{(1-\bar{\alpha}_{t})} \epsilon$ with the time step $t$ found by QBM, the distributions closely follow a Gaussian distribution. That is, our method can effectively identify the time step $t$ at which the approximate input resembles a Gaussian distribution, which is Criteria 1 mentioned in \cref{sec:intiial_t}.

\section{Conclusion}
\label{Conclusion}
In this paper, we propose a novel acceleration framework for zero-shot diffusion-based image restoration. Specifically, we introduce the Quick Bypass Mechanism (QBM) to initialize the denoising process from an intermediate approximation rather than pure noise. Additionally, we present the Revised Reverse Process (RRP) to enhance stochasticity, compensating for the disharmony caused by the approximation. We evaluate our methods on ImageNet-1K and CelebA-HQ across super-resolution, deblurring, and compressed sensing tasks. Our results demonstrate that the proposed method achieves comparable performance to 100-step DDNM while only running 5$\sim$59 steps, depending on the task.

\bibliographystyle{IEEEbib}
\bibliography{strings,refs}

\end{document}